\documentclass[10pt,twocolumn,letterpaper]{article}

\usepackage{cvpr}              
\usepackage[accsupp]{axessibility}  
\newcommand{\xmark}{\ding{55}}
\usepackage{pifont}

\definecolor{cvprblue}{rgb}{0.21,0.49,0.74}
\usepackage[pagebackref,breaklinks,colorlinks,allcolors=cvprblue]{hyperref}

\def\paperID{*****} 
\def\confName{CVPR}
\def\confYear{2025}

\title{Prototype-Based Continual Learning with Label-free Replay Buffer and Cluster Preservation Loss}
\author{Agil Aghasanli, Yi Li, Plamen Angelov\\
School of Computing and Communications, Lancaster University\\\{a.aghasanli1, y.li154, p.angelov\}@lancaster.ac.uk}

\begin{document}
\maketitle
\begin{abstract}
Continual learning techniques employ simple replay sample selection processes and use them during subsequent tasks. Typically, they rely on labeled data. In this paper, we depart from this by automatically selecting prototypes stored without labels, preserving cluster structures in the latent space across tasks. By eliminating label dependence in the replay buffer and introducing cluster preservation loss, it is demonstrated that the proposed method can maintain essential information from previously encountered tasks while ensuring adaptation to new tasks. "Push-away" and "pull-toward" mechanisms over previously learned prototypes are also introduced for class-incremental and domain-incremental scenarios. These mechanisms ensure the retention of previously learned information as well as adaptation to new classes or domain shifts. The proposed method is evaluated on several benchmarks, including SplitCIFAR100, SplitImageNet32, SplitTinyImageNet, and SplitCaltech256 for class-incremental, as well as R-MNIST and CORe50 for domain-incremental setting using pre-extracted DINOv2 features. Experimental results indicate that the label-free replay-based technique outperforms state-of-the-art continual learning methods and, in some cases, even surpasses offline learning. An unsupervised variant of the proposed technique for the class-incremental setting, avoiding labels use even on incoming data, also demonstrated competitive performance, outperforming particular supervised baselines in some cases. These findings underscore the effectiveness of the proposed framework in retaining prior information and facilitating continual adaptation.
\end{abstract}    
\section{Introduction}
\label{introduction}

The increasing demand for intelligent systems that operate in dynamically changing environments requires continuous learning, a paradigm in which models learn and improve over time without losing previous understanding \cite{Parisi2018ContinualLL}. Unlike classic machine learning algorithms, which assume a fixed dataset and a single training session, continuous learning is more closely aligned with real-life situations in which data streams evolve and distributions shift. 

Catastrophic forgetting is a crucial obstacle to achieving such adaptable systems, which was first described in \cite{MCCLOSKEY1989109} showing that sequentially updating connectionist models on new data can overwrite previously learned representations. Subsequent works \cite{Ratcliff1990, FRENCH1999128} also supported this view, mentioning that the interference between old and new representations is usually the reason for the rapid decrease in performance. In larger networks, those inferences may be caused by a phenomenon where many neurons are repurposed during updates, leading to fast and unintentional loss of previously learned representations \cite{Goodfellow2013AnEI}. 

The adverse effects of catastrophic forgetting are particularly significant in cases that need continuous reliability, specifically in the applications of autonomous systems \cite{Shaheen2021ContinualLF}, adaptive user interfaces \cite{Mei2024TrainOnRequestAO}, and robotics \cite{Lesort2019ContinualLF}, which illustrate the need for accumulating knowledge without repeated resets or the luxury of training from scratch \cite{THRUN199525}. If a system suddenly forgets how to perform an older yet crucial task, it could result in safety risks and increase operating costs \cite{Kirkpatrick2016OvercomingCF}. 


In this work, we propose a novel Continual Learning (CL) framework that unifies three key contributions to mitigate catastrophic forgetting and enhance knowledge transfer:

\begin{itemize}
    \item \textbf{Cluster Preservation Loss:} 
    We introduce a loss function that maintains the structure of previously learned clusters by minimizing the effect of the distribution shifts from new tasks over time, thereby preserving critical information from earlier tasks.

    \item \textbf{Push-Away and Pull-Toward Mechanisms:} 
    Tailored for Class-Incremental (CI) and Domain-Incremental (DI) scenarios, respectively, these mechanisms ensure class separation (Push-Away) and domain consistency (Pull-Toward). By segregating tasks into well-separated or well-aligned representations, the model can accommodate new information without overwriting old knowledge.

    \item \textbf{Label-free Replay Buffer:} 
    We store historical samples—represented as class prototypes and support samples—without any label metadata. This approach provides a privacy-preserving alternative to replay methods that depend on labeled examples.
\end{itemize}

We call this methodology \textbf{iSL-LRCP} (\textbf{i}ncremental \textbf{S}upervised \textbf{L}earning with \textbf{L}abel-free \textbf{R}eplay buffer and \textbf{C}luster \textbf{P}reservation) and \textbf{iUL-LRCP} (\textbf{i}ncremental \textbf{U}nsupervised \textbf{L}earning with \textbf{L}abel-free \textbf{R}eplay buffer and \textbf{C}luster \textbf{P}reservation), representing the supervised and unsupervised variants of our framework, respectively. We evaluate our method on both class-incremental (SplitCIFAR100 \cite{cifar100}, SplitImageNet32 \cite{Chrabaszcz2017ADV}, SplitTinyImageNet \cite{Le2015TinyIV}, SplitCaltech101 \cite{caltech256}) and domain-incremental (R-MNIST \cite{deng2012mnist}, CORe50 \cite{pmlr-v78-lomonaco17a}) benchmarks. Unlike many existing techniques that store explicit labels or integrate softmax-based classification layers, our approach employs a nearest prototype classification layer, leveraging the retained prototypes to classify incoming data. Extensive experimental results demonstrate that our Label-Free Replay-Based framework effectively reduces catastrophic forgetting, consistently outperforming strong baselines, including replay-free PRD \cite{Asadi2023PrototypeSampleRD}, replay-based ER-AML \cite{DBLP:conf/iclr/CacciaAATPB22} and iCaRL \cite{Rebuffi2016iCaRLIC}, and even, in some cases, offline learning. Overall, our method provides a unified solution that robustly preserves prior knowledge and flexibly adapts to new data distributions.

\begin{figure*}[h]
\centering
\includegraphics[scale=0.45]{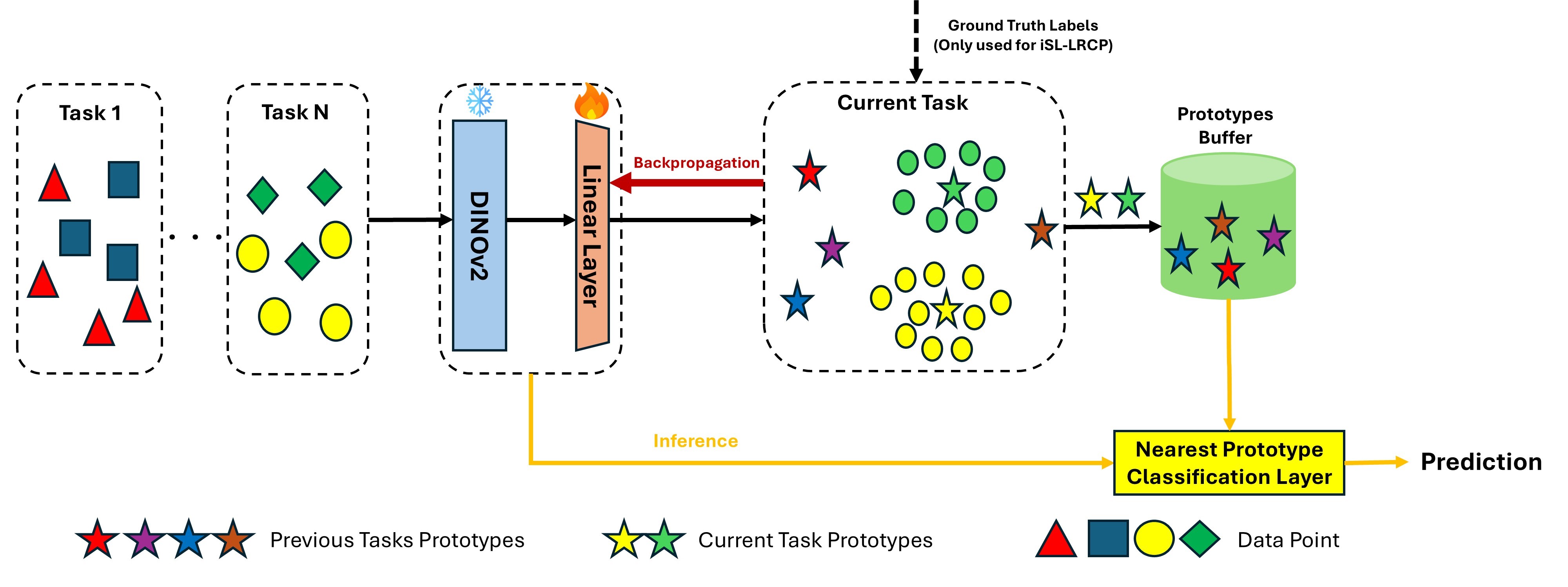}
\caption{Overview of the continual learning framework. Features are extracted using a pre-trained DINOv2 ViT-L/14 backbone and projected into a 512-dimensional space. Class prototypes are computed with K-means and support samples are selected via $\sigma$-band sampling, and then stored in a label-free replay buffer.}
\label{figpipe}
\end{figure*}
\section{Related Work}
\subsection{Continual Learning}
Recent CL approaches have made significant strides in mitigating catastrophic forgetting. These approaches can be broadly categorized into regularization-based methods that constrain weight updates \cite{Kirkpatrick2016OvercomingCF}, replay-based techniques that maintain exemplars of previous tasks \cite{AMR}, parameter isolation strategies that allocate specific sub-networks for different tasks \cite{PI}, and architectural methods that dynamically expand the network capacity \cite{DAA}. As a recent and effective replay-based technique, experience Replay Asymmetric Cross-Entropy (ER-ACE) addresses \cite{DBLP:conf/iclr/CacciaAATPB22} the challenge of representational changes in observed data when previously unseen classes are introduced into the data stream, requiring distinction between new and old classes. Traditional experience replay often results in significant overlap between the representations of the newly added classes and existing ones, causing disruptive parameter updates and leading to catastrophic forgetting. ER-ACE mitigates this by employing an asymmetric update rule, wherein new classes are encouraged to adapt to the representations of older classes rather than the reverse. However, despite recent advances in CL, a fundamental trade-off remains between preserving knowledge of previously learned tasks and efficiently adapting to new ones.

CL scenarios are often categorized into Class-Incremental (CI) learning and Domain-Incremental (DI) learning, each addressing distinct challenges. CI learning requires a model to sequentially learn new classes while retaining knowledge of previously learned ones, demanding a balance between distinguishing new and old classes to avoid catastrophic forgetting. Replay-based or regularization methods are typically employed to mitigate representational overlap between classes for this setting. For example, iCaRL \cite{Rebuffi2016iCaRLIC} selects and stores representative samples utilizing herding techniques, uses a nearest-mean-of-exemplars classifier with knowledge distillation to enable CI learning and prevent catastrophic forgetting. On the other hand, DI learning focuses on adapting to changes in input data distribution across tasks requiring robust strategies to generalize the knowledge and apply to diverse domains. Specifically, EWC \cite{Kirkpatrick2016OvercomingCF} works by adding a regularization term to the loss function that penalizes changes to weights critical for previously learned tasks based on their importance estimated using the Fisher Information Matrix, and this approach has been effectively demonstrated on the Permuted MNIST \cite{deng2012mnist} dataset to mitigate catastrophic forgetting in CL settings. Both settings highlight the trade-offs inherent in CL methods, particularly regarding stability-plasticity and computational efficiency in memory-constrained environments.

\subsection{Prototype Learning}
Class prototypes represent an essential concept in machine learning that captures the archetypal characteristics of different categories through learned representations \cite{Bien2011PrototypeSF}. This involves learning centroids or exemplars in a feature space that embody the fundamental properties shared among instances of the same class, enabling more interpretable and robust classification systems \cite{ANGELOV2025129464}. By distilling complex data distributions into representative prototypes, these methods create intuitive decision boundaries while maintaining competitive performance with traditional approaches.

Recent advances in prototype-based learning have shown remarkable effectiveness across various applications, including cyber security \cite{eccv}, continual learning \cite{Asadi2023PrototypeSampleRD}, and deepfake detection \cite{cvprw}. Beyond their practical utility, prototypes offer a bridge between instance-based and parametric learning approaches, making them particularly valuable for scenarios requiring both model interpretability and performance. For example, \cite{Asadi2023PrototypeSampleRD} proposed a comprehensive approach to prototype extraction that integrates representation and class information. This is achieved by mapping samples into a feature space using supervised contrastive loss. Class prototypes are being continually updated within the same latent space, enabling both learning and prediction. The method ensures that class prototypes retain their relative similarities to new task data while dynamically adapting, eliminating the need to store data from prior tasks. 
\section{Methodology}
The proposed methodology addresses the challenges of continual learning in both CI and DI settings. It integrates prototype-based classification and novel loss functions to achieve information retention and adaptation to new tasks. The framework is presented in Fig. \ref{figpipe}.

\subsection{Model Architecture and Feature Extraction}\label{subsection:architecture}
Firstly, the proposed framework utilizes features from the pre-trained DINOv2 ViT-L/14 architecture \cite{Oquab2023DINOv2LR}, represented as 1024 dimensional vectors. Then, these features are transformed into a 512 dimensional latent space using a linear layer, where the linear model is trained incrementally on new tasks. This architecture enables task-specific representation learning while supporting prototype-based classification.

\subsection{Prototype Computation and Buffer Storage}

Once the features are transformed into the latent space after training on each task with one of the combined loss functions described in Section \ref{subsection:losses}, The class prototypes and support samples are determined to represent task-specific and cluster-wise information.

\paragraph{Class Prototypes and Support Samples:} Class prototypes and support samples are crucial components in our methodology, designed to preserve information across tasks and enhance model evaluation. Class prototypes are determined as the nearest samples to the center of each cluster, which are computed using K-means clustering. The number of clusters (\(N\)) is set equal to the number of classes (\(K\)) in each task $N=K$, ensuring that each cluster is represented by a single, meaningful prototype. In addition, support samples are selected to capture the spread of each cluster by identifying samples along specific sigma bands, such as $\pm 1\sigma$, $\pm 2\sigma$, and $\pm 3\sigma$.

\paragraph{Support Sample Selection.} To ensure that support samples capture the each cluster's distribution, we first perform dimension selection identifying the most informative and non-redundant dimensions based on the principles of the minimum Redundancy Maximum Relevance (mRMR) method \cite{1453511}. However, unlike that technique, an unsupervised variance-based relevance selection strategy is employed in our paper.

The dimension selection process in this work is consisted of three steps. First, the variance per dimension within a cluster is computed to assess the informativeness. Next, dimensions with the highest variance are prioritized. Finally, those with correlation above a threshold \( \epsilon \) with already selected dimensions are discarded to reduce redundancy.

Once informative features are selected as \( d \), we determine target points at \( \mu \pm k \sigma_d \) ($k \in \{1,2,3\}$) and identify the nearest samples to these points as support samples. This ensures the support samples effectively capture the spread of each cluster while preserving structural integrity.

The class prototypes and support samples are stored in a label-free replay buffer, which includes both the input representations and their corresponding latent representations. This buffer serves as a compact memory, facilitating knowledge transfer and retrieval during subsequent tasks. 

\subsection{Loss Functions}\label{subsection:losses}

\paragraph{\textbf{Supervised Contrastive Loss:}} 

We employ the Supervised Contrastive Loss as outlined in \cite{NEURIPS2020_d89a66c7}. This loss aligns same-class samples more closely in the latent space while increasing separation between different-class samples:

\begin{equation}
\mathcal{L}_{\text{sc}} = \frac{1}{N} \sum_{i=1}^{N} \frac{-1}{|P(i)|} \sum_{p \in P(i)} \log \frac{\exp(\mathbf{z}_i \cdot \mathbf{z}_p / \tau)}{\sum_{a \in A(i)} \exp(\mathbf{z}_i \cdot \mathbf{z}_a / \tau)},
\end{equation}

where \( N \) is the batch size, \( \mathbf{z}_i \) represents the normalized latent representation of sample \( i \), \( P(i) \) is the set of positive samples sharing the same label, \( A(i) \) is the set of all other samples, and \( \tau \) is the temperature parameter controlling similarity scaling. This loss enhances class discriminability in both class-incremental (CI) and domain-incremental (DI) settings.

\paragraph{\textbf{Cluster Preservation Loss:}} 

In this paper, we introduce the Cluster Preservation loss, which is designed to maintain the structural integrity of previously learned clusters during incremental learning. By leveraging the Maximum Mean Discrepancy (MMD) metric \cite{JMLR:v13:gretton12a}, it measures and minimizes the effect of the distributional shift from the new task samples. Our novelty lies in applying this metric to class prototypes and support samples, using their latent representations before (\( \mathbf{Z}_{\text{old}} \)) and after (\( \mathbf{Z}_{\text{new}} \)) training on a new batch of task samples. The loss is expressed as:

\begin{equation}
\mathcal{L}_{\text{preserve}} = \text{MMD}^2\left( \mathbf{Z}_{\text{old}}, \mathbf{Z}_{\text{new}} \right),
\end{equation}

This ensures that cluster structures remain consistent over time, preventing catastrophic forgetting by preserving their representation in the latent space.

\begin{figure}[h]
\centering
\includegraphics[scale=0.3]{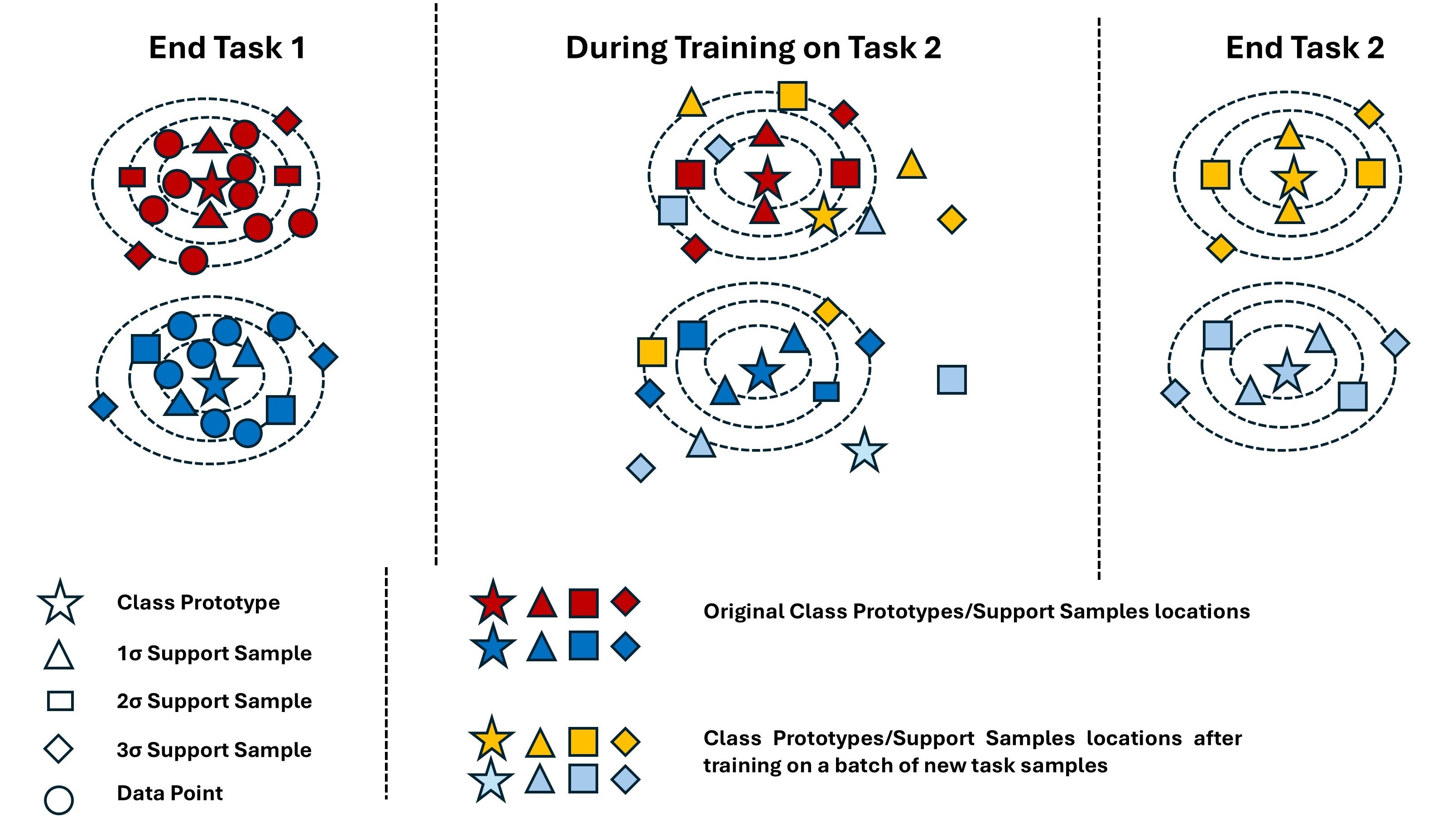}
\caption{Illustration of the Cluster Preservation Loss. This mechanism retains the structural integrity of clusters by minimizing distribution shifts of class prototypes and support samples in the latent space across tasks. Dashed lines indicate the $\sigma$-bands within which support samples are selected, ensuring consistency and preventing catastrophic forgetting.}
\label{fig_cluster_preserve}
\end{figure}

\paragraph{\textbf{Contrastive Push-Away Loss:}} 

The Contrastive Push-Away Loss, introduced as a novel component of this work, ensures that the representations of new tasks remain distinct from those of previously learned classes. This is achieved by penalizing the excessive similarity between the new representations and the prototypes of prior classes. The loss is formally expressed as:

\begin{equation}
\mathcal{L}_{\text{push}} = \frac{1}{N} \sum_{i=1}^{N} \sum_{j=1}^{C_{\text{prev}}} \frac{\mathbf{z}_i \cdot \boldsymbol{\mu}_j}{(1 - \sigma_j) \cdot \tau_{push}} ,
\end{equation}

where \( \mathcal{L}_{\text{push}} \) calculates the average similarity between the current sample \( \mathbf{z}_i \) and the prototype \( \boldsymbol{\mu}_j \) of previously learned classes. Here, \( \mathbf{z}_i \) is the latent representation of the \( i^{th} \) sample, while \( \boldsymbol{\mu}_j \) represents the mean of the latent representations for the \( j^{th} \) previous class. The standard deviation \( \sigma_j \) reflects the spread of representations for the \( j^{th} \) class, and the term \( (1 - \sigma_j) \) inversely weights similarity based on the compactness of the class cluster. \( C_{\text{prev}} \) denotes the number of classes from earlier tasks, and \( \tau_{push} \) is the temperature parameter used to scale similarity scores. By minimizing this loss, the model ensures sufficient separation between the new task representations and previously learned class prototypes, promoting distinct and non-overlapping clusters in the latent space.

The inverse dependence in regards to \(1 - \sigma_j\) places greater emphasis on separating representations from loosely packed clusters (higher \(\sigma_j\)), where distinguishing boundaries might be less clear due to their natural dispersion. In contrast, tightly packed clusters (lower \(\sigma_j\)) receive less emphasis, as their compactness inherently aids separability.

\begin{figure}[h]
\centering
\includegraphics[scale=0.3]{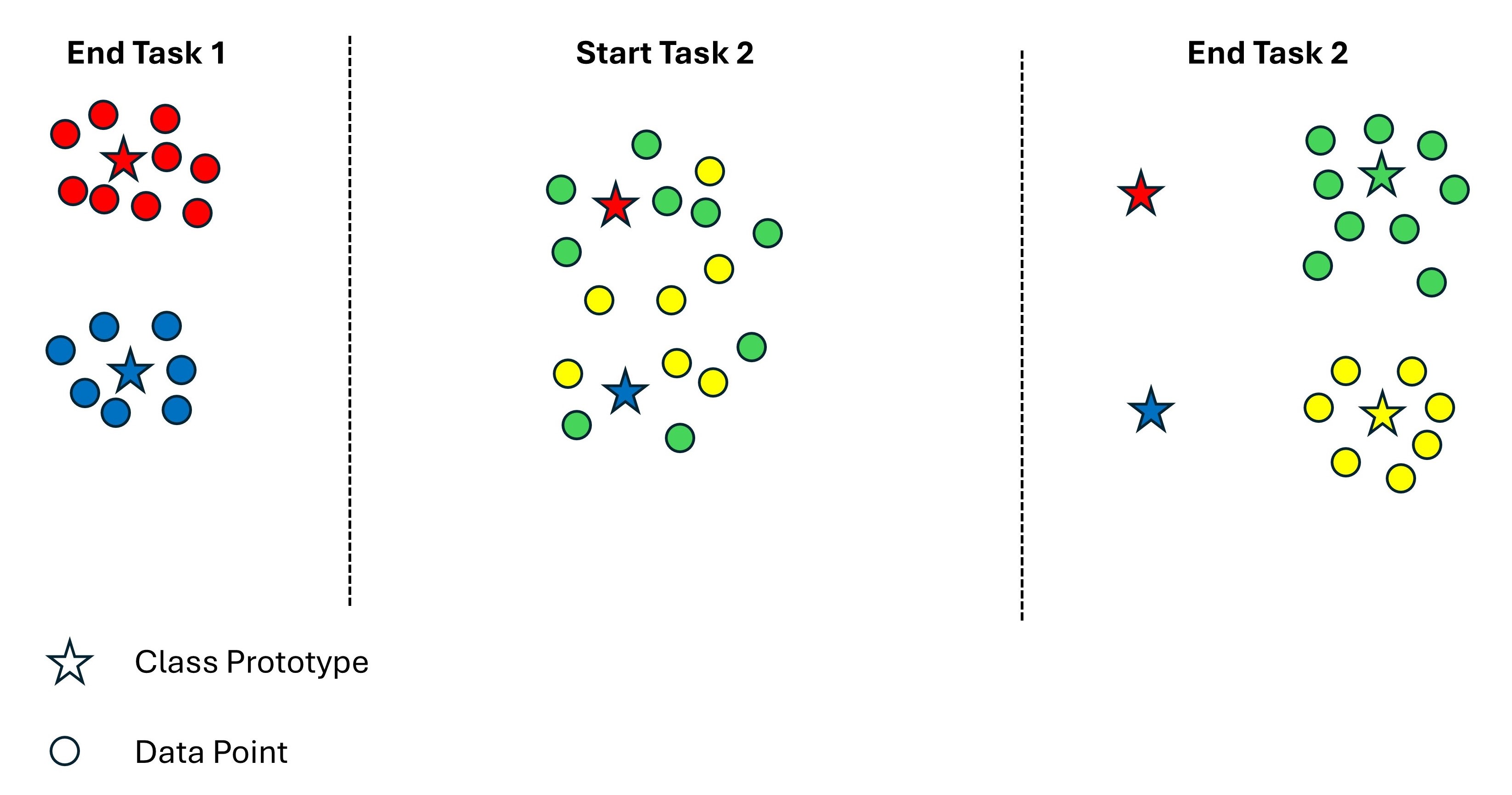}
\caption{The combined Supervised Contrastive and Push-Away Loss. The supervised contrastive loss ensures intra-class compactness and inter-class separation, while the push-away loss enforces additional separation between latent representations of new data points and prototypes from previously learned tasks, mitigating interference.}
\label{fig_push_away}
\end{figure}

\paragraph{\textbf{Contrastive Pull-Toward Loss:}}

The Contrastive Pull-Toward Loss, also newly introduced in this work, aligns current sample representations with prototypes from the first task, promoting domain-invariant features and reducing domain shifts. It is defined as:

\begin{equation}
\mathcal{L}_{\text{pull}} = \frac{1}{N} \sum_{i=1}^{N} \sum_{j=1}^{C_{\text{prev}}} \delta(y_i, y_{\mu_j}) \left( 1 - \cos(\mathbf{z}_i, \boldsymbol{\mu}_j) \right),
\end{equation}

where \( \mathbf{z}_i \) and \( \boldsymbol{\mu}_j \) are the normalized representations of the sample and the first task prototype, respectively. The indicator function \( \delta(y_i, y_{\mu_j}) \) ensures alignment only for matching classes. Minimizing this loss reduces angular distance, mitigating domain shifts, and improving DI performance.

\begin{figure}[h]
\centering
\includegraphics[scale=0.28]{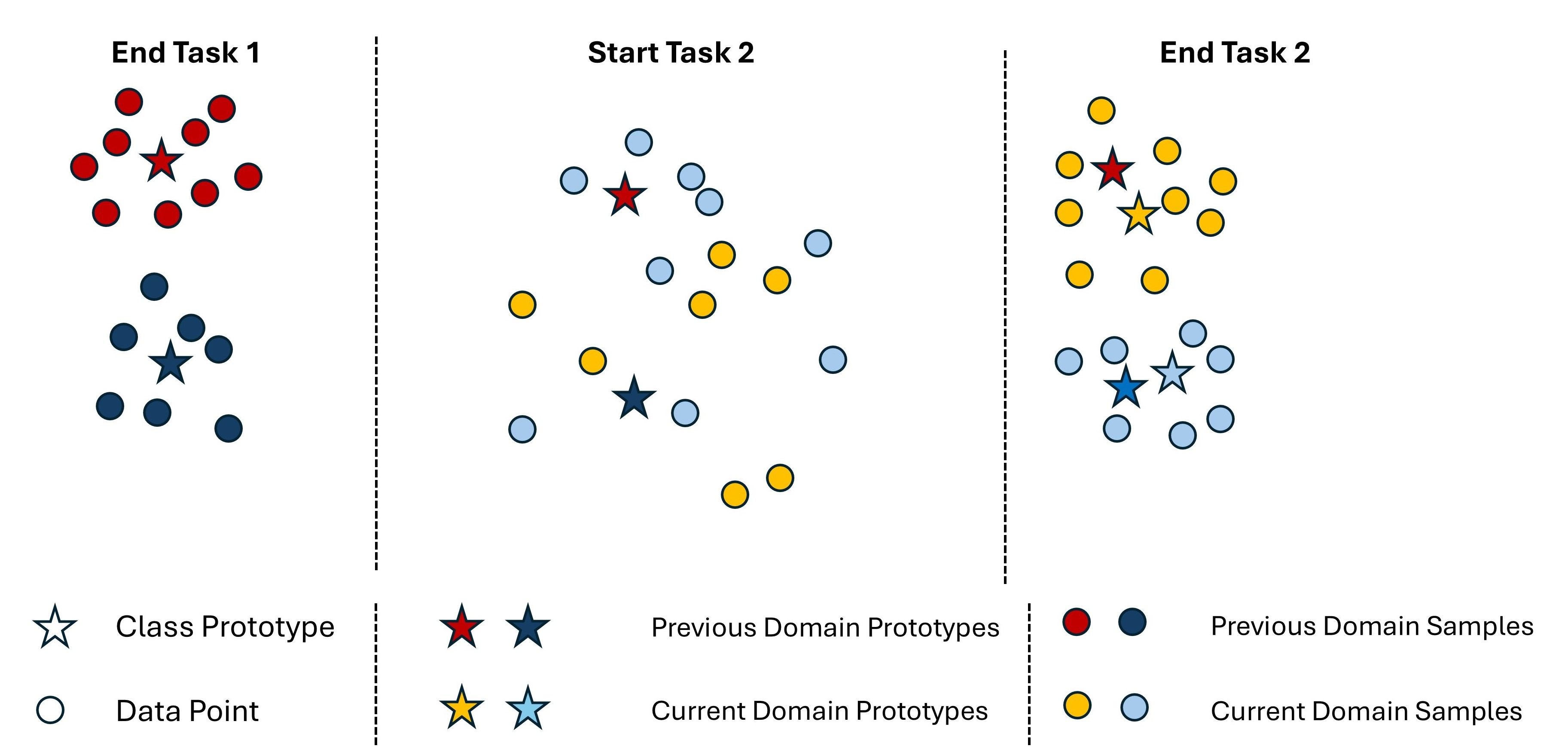}
\caption{Illustration of the combined Supervised Contrastive and Pull-Toward Loss. The supervised contrastive loss ensures intra-class compactness and inter-class separation within the current domain, while the pull-toward loss aligns representations of new domain samples with prototypes from the first domain, promoting consistency across domains and reducing domain shifts.}
\label{fig_pull_to}
\end{figure}

\paragraph{\textbf{Pseudo-Contrastive Loss:}}

The Pseudo-Contrastive Loss, proposed in this work for unsupervised continual learning, dynamically assigns pseudo-labels using MiniBatch K-means clustering on latent representations of new batches of data samples. The loss is defined as:

\begin{equation}
\mathcal{L}_{\text{pc}} = \frac{1}{N} \sum_{i=1}^{N} \frac{-1}{|P'(i)|} \sum_{p \in P'(i)} \log \frac{\exp(\mathbf{z}_i \cdot \mathbf{z}_p / \tau)}{\sum_{a \in A(i)} \exp(\mathbf{z}_i \cdot \mathbf{z}_a / \tau)},
\end{equation}

where \( P'(i) \) represents the set of positive samples for \( i \), derived from the pseudo-labels assigned via MiniBatch K-means clustering. \( A(i) \) is the set of all other samples, and \( \tau \) is a temperature parameter. The proposed approach leverages pseudo-labels obtained from clustering to dynamically define \( P'(i) \), enabling effective representation learning without labeled data.

\paragraph{Combined Loss Function.} The total loss function varies for different scenarios:
\begin{itemize}
    \item \textbf{Class-Incremental (Supervised):}
    \begin{equation}
    \mathcal{L}_{\text{total}} = \mathcal{L}_{\text{sc}} + \lambda_{\text{push}} \cdot \mathcal{L}_{\text{push}} + \lambda_{\text{preserve}} \cdot \mathcal{L}_{\text{preserve}}.
    \end{equation}
    \item \textbf{Class-Incremental (Unsupervised):}
    \begin{equation}
    \mathcal{L}_{\text{total}} = \mathcal{L}_{\text{pc}} + \lambda_{\text{push}} \cdot \mathcal{L}_{\text{push}} + \lambda_{\text{preserve}} \cdot \mathcal{L}_{\text{preserve}}.
    \end{equation}
    \item \textbf{Domain-Incremental:}
    \begin{equation}
    \mathcal{L}_{\text{total}} = \mathcal{L}_{\text{sc}} + \lambda_{\text{pull}} \cdot \mathcal{L}_{\text{pull}} + \lambda_{\text{preserve}} \cdot \mathcal{L}_{\text{preserve}}.
    \end{equation}
\end{itemize}


Class prototypes, selected through clustering in the latent space, are utilized during the evaluation phase. The nearest prototype classification mechanism assigns class labels to incoming samples by calculating their distance to the stored class prototypes in the latent space. This approach ensures an interpretable classification without the need for additional softmax-based classifiers.

\section{Experiments}
This section presents the evaluation of the proposed method across CI and DI learning tasks. Detailed analyses of datasets, baselines and results are provided, along with discussions on the findings.

\begin{figure*}[t]
    \centering
    \begin{subfigure}[b]{0.48\textwidth}
        \centering
        \includegraphics[width=\textwidth, height=4cm]{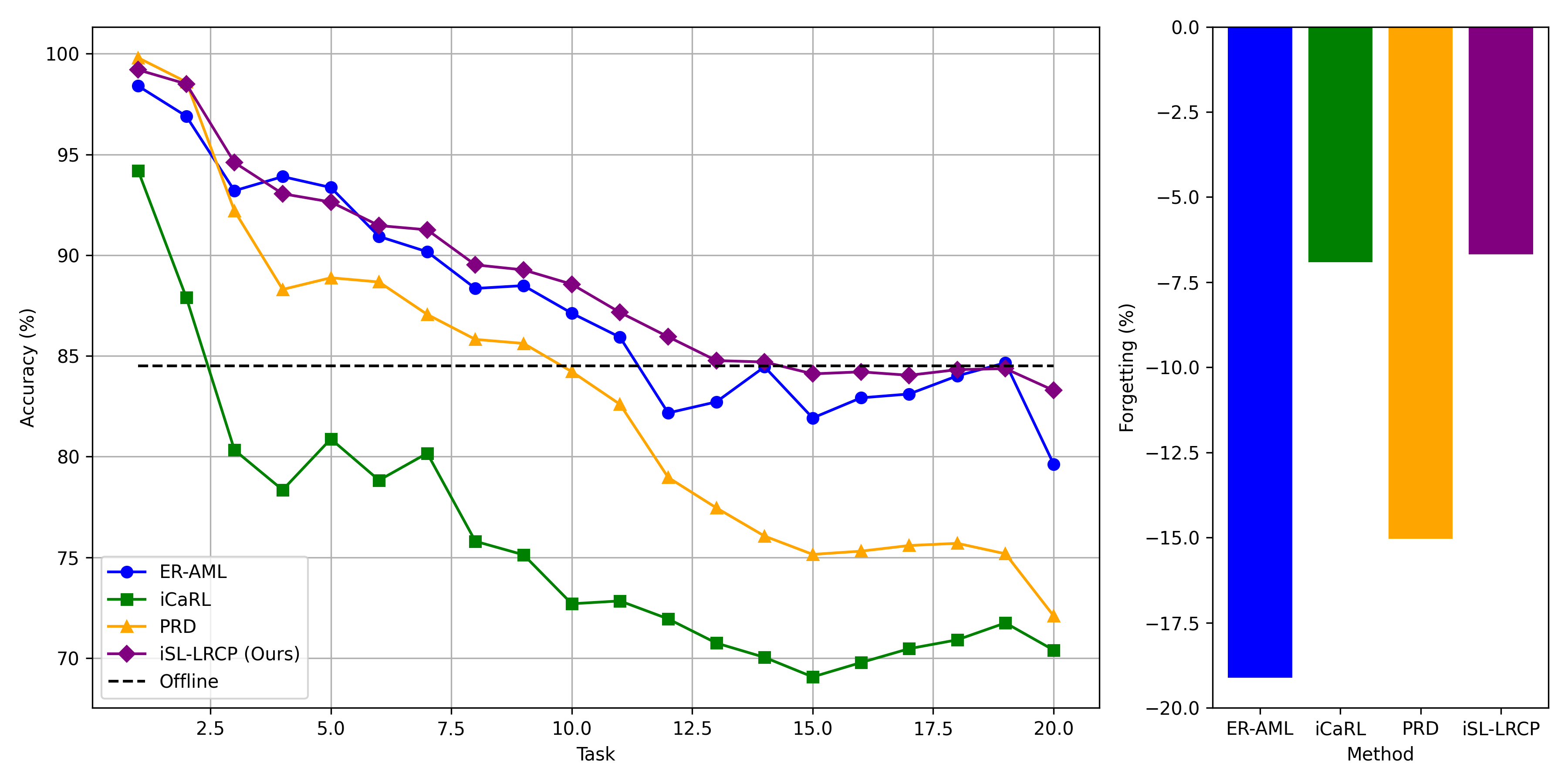}
        \caption{Accuracy and forgetting over 20 tasks for SplitCIFAR100.}
        \label{fig:cifar100_accuracy}
    \end{subfigure}
    \hfill
    \begin{subfigure}[b]{0.48\textwidth}
        \centering
        \includegraphics[width=\textwidth, height=4cm]{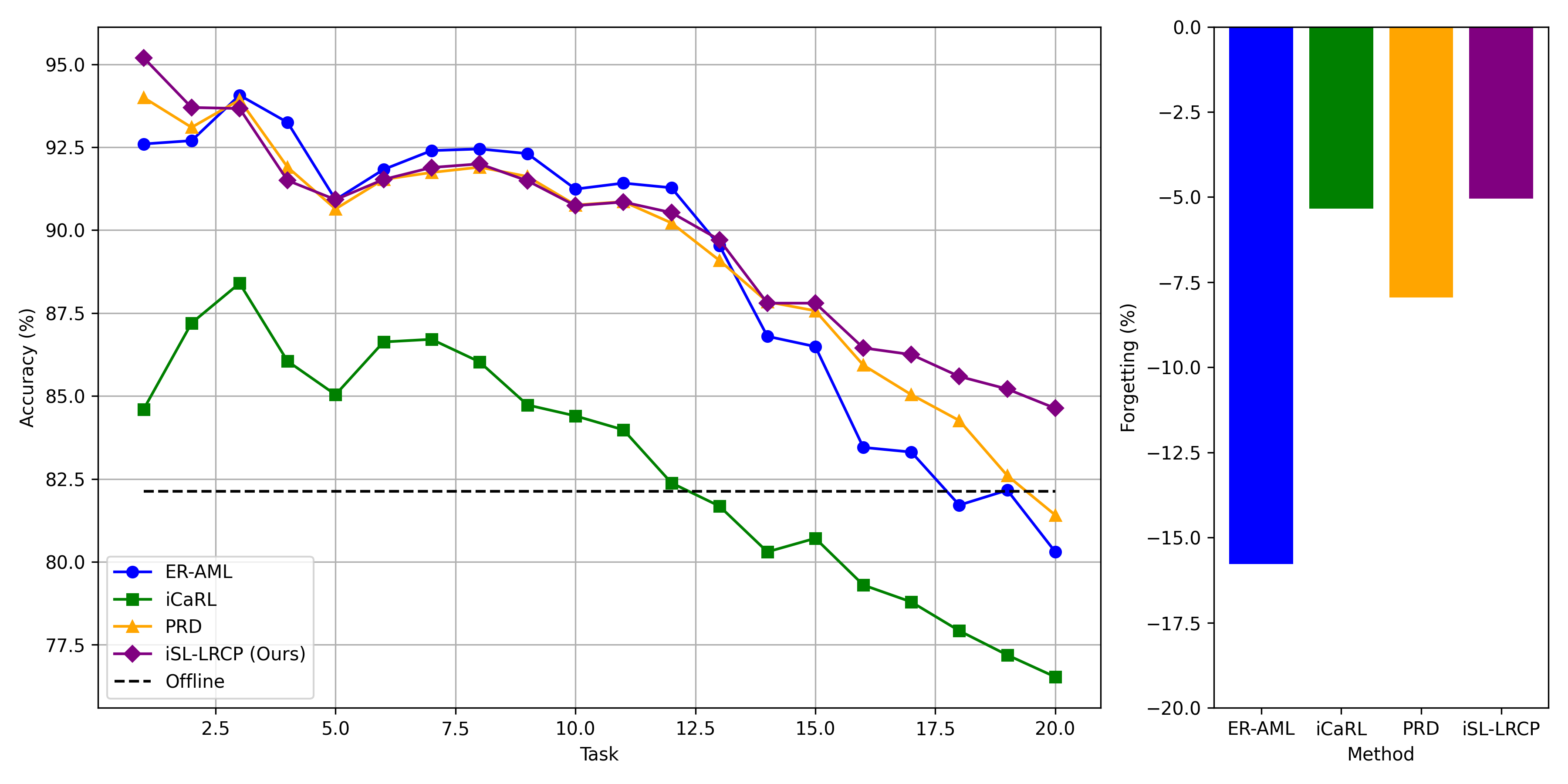}
        \caption{Accuracy and forgetting over 20 tasks for SplitTinyImageNet.}
        \label{fig:tiny_imagenet_accuracy}
    \end{subfigure}
    \caption{Comparison of accuracy and forgetting scores for different methods across CI datasets. (a) SplitCIFAR100; (b) SplitTinyImageNet.}
    \label{fig:accuracy_forgetting}
\end{figure*}

\subsection{Datasets}

Experiments were conducted on several benchmarks for both CI and DI learning settings. Datasets used for CI setting are described in Table \ref{table:class-incremental-datasets}.





\begin{table}[h]
\centering
\caption{Datasets used for the CI setting. The first 500 classes from the ImageNet32 dataset are used for SplitImageNet32.}
\label{table:class-incremental-datasets}
\vskip 0.15in
\renewcommand{\arraystretch}{0.8} 
\begin{tabular}
{p{0.2\textwidth}p{0.1\textwidth}p{0.1\textwidth}}
\hline
\textbf{Dataset}         & \textbf{\# Tasks} & \textbf{Class/Task} \\
\hline
SplitCIFAR100            & 20                       & 5                         \\
SplitCaltech256          & 16                       & 16                        \\
SplitTinyImageNet        & 20                       & 10                        \\
SplitImageNet32          & 50                       & 10                        \\
\hline
\end{tabular}
\vskip -0.1in
\end{table}

For the DI setting, two datasets are used. First, six distinct tasks are generated by applying rotations of $60^\circ$, $120^\circ$, $180^\circ$, $240^\circ$, and $300^\circ$ to 10,000 images from the MNIST \cite{deng2012mnist} (1,000 images per class for 10 classes) to obtain the R-MNIST dataset. Second, 11 tasks are used based on object categories under varying conditions in the CORe50 dataset \cite{pmlr-v78-lomonaco17a}.

\subsection{Baselines}

To evaluate the performance of the proposed method, it is compared against the following baselines:

\textbf{ER-AML} \cite{DBLP:conf/iclr/CacciaAATPB22}: This method employs an asymmetric metric learning loss to mitigate representation drift during replay. It showed competitive results among replay-based methods and outperformed baselines such as ER \cite{chaudhry2019tiny}, DER++ \cite{10.5555/3495724.3497059}, and SS-IL \cite{Ahn2020SSILSS}.

\textbf{iCaRL} \cite{Rebuffi2016iCaRLIC}: A strong replay-based baseline that selects the mean of embeddings for each class as prototypes and uses nearest prototype classification for evaluation.

\textbf{PRD} \cite{Asadi2023PrototypeSampleRD}: A replay-free CL method that evolves class prototypes in the latent space using a prototype-sample relation distillation loss. It demonstrated state-of-the-art results among replay-free methods such as LwF \cite{Li2016LearningWF}, EWC \cite{Kirkpatrick2016OvercomingCF}, and SPB \cite{Wu2021}.

\textbf{Offline Baseline}: The offline baseline trains the model with supervised contrastive loss on the entire dataset. Prototypes are identified using K-means clustering, and nearest prototype classification is used for evaluation, same as the proposed method.

We evaluated both supervised iSL-LRCP and unsupervised iUL-LRCP variants of the proposed method. This comprehensive evaluation provides information on the robustness and adaptability of the approach across various CL settings.

\begin{table*}[t]
\centering
\caption{Performance comparison of different methods on the CI setting across various datasets. The datasets are abbreviated as \textbf{sC100} (SplitCIFAR100, 20 tasks), \textbf{sCal256} (SplitCaltech256, 16 tasks), \textbf{sTinyIN} (SplitTinyImageNet, 20 tasks), and \textbf{sIN32} (SplitImageNet32, 50 tasks). \textbf{*} Indicates the proposed method outperformed the offline baseline. The underlined results represent supervised baselines that were outperformed by iUL-LRCP.}
\label{comparison-table-class-inc}
\vskip 0.15in
\renewcommand{\arraystretch}{0.9} 
\begin{small}
\begin{tabular}{lccc cccc}
\toprule
\textbf{Method} & \textbf{Supervised} & \textbf{Replay Buffer} & \textbf{Label-free} & \textbf{sC100} & \textbf{sCal256} & \textbf{sTinyIN} & \textbf{sIN32} \\
& & & \textbf{Replay Buffer} & \textbf{(20 Tasks)} & \textbf{(16 Tasks)} & \textbf{(20 Tasks)} & \textbf{(50 Tasks)} \\
\midrule
Offline       & \checkmark & -      & -      & 84.50\% & 92.00\% & 82.13\% & 60.25\% \\
\midrule
iSL-LRCP (Ours)        & \checkmark & \checkmark & \checkmark & \textbf{83.29\%} & \textbf{92.87\%}$^*$ & \textbf{84.63\%}$^*$ & \textbf{58.67\%} \\
ER-AML & \checkmark & \checkmark & \xmark & 79.63\% & 92.20\% & 80.30\% & \underline{25.57\%} \\
PRD    & \checkmark & \xmark & -      & \underline{72.10\%} & \underline{29.17\%} & 81.91\% & 55.54\% \\
iCaRL & \checkmark & \checkmark & \xmark & \underline{70.40\%} & 89.02\% & 76.53\% & 58.02\% \\
\midrule
iUL-LRCP (Ours) & \xmark & \checkmark & \checkmark & 77.08\% & 83.48\% & 74.14\% & 51.92\% \\
\bottomrule
\end{tabular}
\end{small}
\vskip -0.1in
\end{table*}

\begin{table}[t]
\centering
\caption{Backward Transfer (BWT) results showing the degree of forgetting across different methods and datasets.}
\label{table-bwt-results}
\vskip 0.15in
\renewcommand{\arraystretch}{1.1} 
\begin{center}
\begin{footnotesize}
\begin{tabular}{p{0.13\textwidth}p{0.06\textwidth}p{0.06\textwidth}p{0.06\textwidth}p{0.06\textwidth}}
\hline
\textbf{Method} & \textbf{sC100} & \textbf{sCal256} & \textbf{sTinyIN} & \textbf{sIN32}  \\ 
\hline
iSL-LRCP (Ours) & \textbf{-6.68\%}  & \textbf{-2.30\%} & \textbf{-5.05\%} & -12.77\% \\ 
iCaRL & -6.91\% & -3.46\% & -5.34\% & \textbf{-7.06\%}  \\ 
PRD & -15.03\% & -15.71\% & -7.95\% & -27.78\%  \\ 
ER-AML & -19.12\% & -5.22\% & -15.78\% & -65.68\% \\ 
\hline
\end{tabular}
\end{footnotesize}
\end{center}
\vskip -0.1in
\end{table}

\begin{table}[t]
\centering
\caption{Domain-incremental comparison of methods showing average accuracy on R-MNIST and CORe50 datasets.}
\label{comparison-table-domain-inc}
\vskip 0.15in
\renewcommand{\arraystretch}{1.0} 
\begin{center}
\begin{footnotesize}
\textbf{*} Indicates the proposed method outperformed the offline baseline.
\end{footnotesize}
\begin{tabular}{p{0.19\textwidth}p{0.09\textwidth}p{0.09\textwidth}}
\hline
\textbf{Method} & \textbf{R-MNIST  (6 Tasks)} & \textbf{CORe50 (11 Tasks)}  \\ 
\hline
Offline & 90.00\% & 95.33\% \\ 
\hline
iSL-LRCP (Ours) & \textbf{87.55\%} & \textbf{97.71\%*} \\ 
ER-AML & 71.52\% & 91.10\% \\ 
iCaRL & 53.45\% & 70.17\%  \\ 
\hline
\end{tabular}
\end{center}
\vskip -0.1in
\end{table}

\subsection{Implementation Details}

The experiments were conducted using a machine equipped with an NVIDIA GeForce RTX 3080 GPU. The linear model described in Section \ref{subsection:architecture} was optimized using the Adam optimizer with a learning rate of $1 \times 10^{-4}$. The same linear layer was trained with all baselines and the proposed method.

We ensured consistent hyperparameter settings across all methods to facilitate fair comparisons. For replay-based baselines and the proposed method, the buffer size per class was fixed at $M=31$. Each model was trained with a batch size of 64 over 5 epochs. 

For the ER-AML baseline, a temperature of \( \tau = 0.07 \) was employed for the supervised contrastive loss, optimizing the representation learning process. Similarly, for the iCaRL method, the temperature parameter for the distillation loss was set to \( \tau = 2 \), ensuring knowledge distillation from previous tasks. The PRD method utilized several hyperparameters, including \( \tau_{\text{supcon}} = 0.1 \) for supervised contrastive loss, \( \beta_{\text{distill}} = 4.0 \) to weight the distillation loss, \( \alpha_{\text{prototypes}} = 2.0 \) and \( \eta_{\text{prototypes}} = 0.01 \) for prototype-related adjustments, and \( \tau_{\text{distill}} = 1.0 \) to balance the distillation process.

For the proposed methods, iSL-LRCP and iUL-LRCP, five selected dimensions (\( d = 5 \)) were used, with a threshold correlation \( \epsilon \) of 0.3 applied to filter out redundant dimensions. The supervised contrastive loss temperature was set to \( \tau = 0.07 \), and the contrastive push-away loss temperature was \( \tau_{\text{push}} = 7 \). Additionally, for the CI setting, the weight of the cluster preservation loss was \( \lambda_{\text{preserve}} = 0.5 \), while the push-away loss was weighted with \( \lambda_{\text{push}} = 2.0 \). In the DI setting, \( \lambda_{\text{preserve}} \) was set to 0.05, and \( \lambda_{\text{pull}} \) was 0.1 to better handle domain shifts.

\subsection{Evaluation Metrics}

The proposed framework is evaluated using two key metrics, which assess the balance between knowledge retention and adaptability:

\paragraph{Average Accuracy.}
This metric reflects the mean accuracy across all tasks after training is completed.

\paragraph{Backward Transfer (BWT).}
BWT \cite{lopez2017gradient} quantifies catastrophic forgetting by evaluating the impact of learning new tasks on previously learned tasks:
\begin{equation}
\text{BWT} = \frac{1}{t-1} \sum_{i=1}^{t-1} \left( A_{i,t} - A_{i,i} \right),
\end{equation}
where $A_{i,t}$ is the accuracy on task $i$ after training on task $t$, and $A_{i,i}$ is the accuracy on task $i$ immediately after training on it.

\subsection{Class-Incremental Setting Results}

The performance of the proposed method and baseline approaches across CI learning tasks is summarized in Table \ref{comparison-table-class-inc}. The proposed method consistently outperformed all CL baselines across all datasets, highlighting its effectiveness in balancing adaptability to new tasks with knowledge retention. Notably, the supervised variant of the proposed method iSL-LRCP surpassed the offline baseline on SplitCaltech256 and SplitTinyImageNet datasets, achieving 92.87\% and 84.63\%, respectively.

The iUL-LRCP also demonstrated competitive performance, outperforming supervised baselines such as PRD \cite{Asadi2023PrototypeSampleRD} and iCaRL \cite{Rebuffi2016iCaRLIC} on SplitCIFAR100 (sC100) \cite{cifar100}, PRD on SplitCaltech256 (sCal256) \cite{caltech256}, and ER-AML \cite{DBLP:conf/iclr/CacciaAATPB22} on SplitImageNet32 (sIN32) \cite{Chrabaszcz2017ADV}. This indicates the robustness of the unsupervised approach even without access to labeled data.

Backward Transfer (BWT) results, as shown in Table \ref{table-bwt-results}, further confirm the effectiveness of the proposed method in minimizing forgetting. The proposed method exhibited the lowest negative transfer across most datasets, including sC100 (\(-6.68\%\)), sCal256 (\(-2.30\%\)), and sTinyIN \cite{Le2015TinyIV} (\(-5.05\%\)), significantly outperforming PRD and ER-AML. Although iCaRL achieved slightly lower forgetting on sIN32 (\(-7.06\%\) compared to \(-12.77\%\)), it had substantially lower overall accuracy than the proposed method. These findings demonstrate the ability of the proposed method to achieve both high accuracy and low forgetting, making it an effective solution for CI learning tasks.

\subsection{Domain-Incremental Setting Results}

The results for DI tasks are summarized in Table \ref{comparison-table-domain-inc}, where the 
iSL-LRCP demonstrated superior performance compared to all baselines on both R-MNIST \cite{deng2012mnist} and CORe50 \cite{pmlr-v78-lomonaco17a} datasets. On CORe50, the proposed method achieved an accuracy of 97.71\%, surpassing even the offline baseline (95.33\%), highlighting its ability to effectively handle domain shifts while maintaining high performance. On R-MNIST, it achieved an accuracy of 87.55\%, significantly outperforming ER-AML (71.52\%) and iCaRL (43.45\%), further showcasing its effectiveness in DI learning scenarios. These results underscore the effectiveness of the proposed method in adapting to new domains while preserving knowledge from earlier tasks.

\begin{table}[t]
\centering
\caption{Ablation study evaluating the impact of different loss components on average accuracy across datasets.}
\label{table-loss-comparison}
\vskip 0.15in
\begin{center}
\begin{small}
\renewcommand{\arraystretch}{0.8} 
\setlength{\tabcolsep}{4pt} 
\begin{tabular}{ccc ccc}
\toprule
\multicolumn{3}{c}{\textbf{Loss Function}} & \multicolumn{3}{c}{\textbf{Accuracy}} \\
\cmidrule(lr){1-3} \cmidrule(lr){4-6}
$L_\text{sc}$ & $L_\text{push}/L_\text{pull}$ & $L_\text{preserve}$ & \textbf{sC100} & \textbf{R-MNIST} & \textbf{CORe50} \\
\midrule
\checkmark & \checkmark & \checkmark & \textbf{83.29\%} & \textbf{87.55\%} & \textbf{97.71\%} \\
\checkmark & \xmark & \checkmark & 82.48\% & 87.33\% & 97.03\% \\
\checkmark & \checkmark & \xmark & 20.70\% & 78.27\% &  96.65\% \\
\bottomrule
\end{tabular}
\end{small}
\end{center}
\vskip -0.1in
\end{table}

\begin{figure*}[t]
    \centering
    \begin{subfigure}[b]{0.48\textwidth}
        \centering
        \includegraphics[width=0.75\textwidth, height=2.5cm]{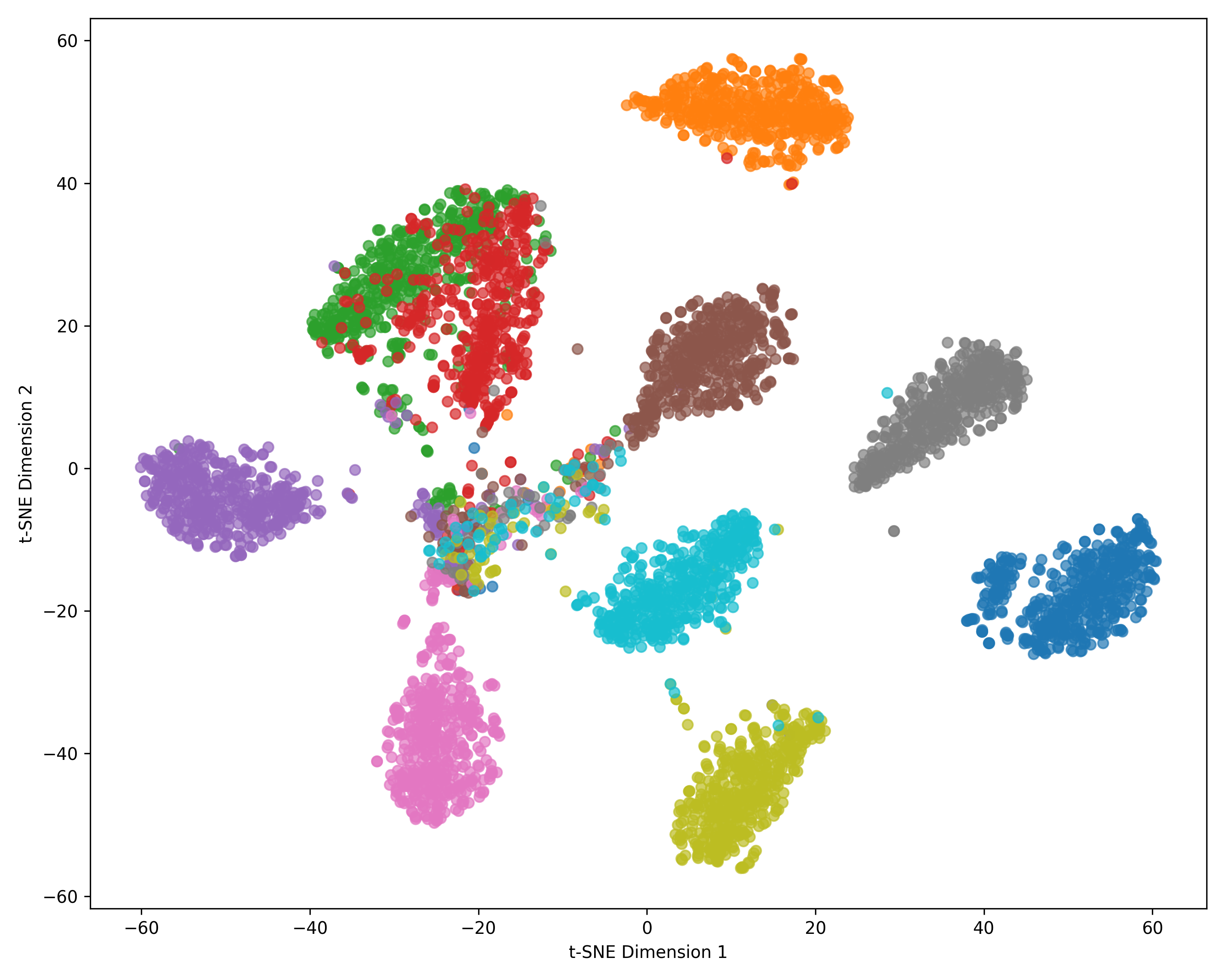}
        \caption{ER-AML: t-SNE visualization after training on task 1.}
        \label{fig:tsne_task1_after_task1_ER-AML}
    \end{subfigure}
    \hfill
    \begin{subfigure}[b]{0.48\textwidth}
        \centering
        \includegraphics[width=0.75\textwidth, height=2.5cm]{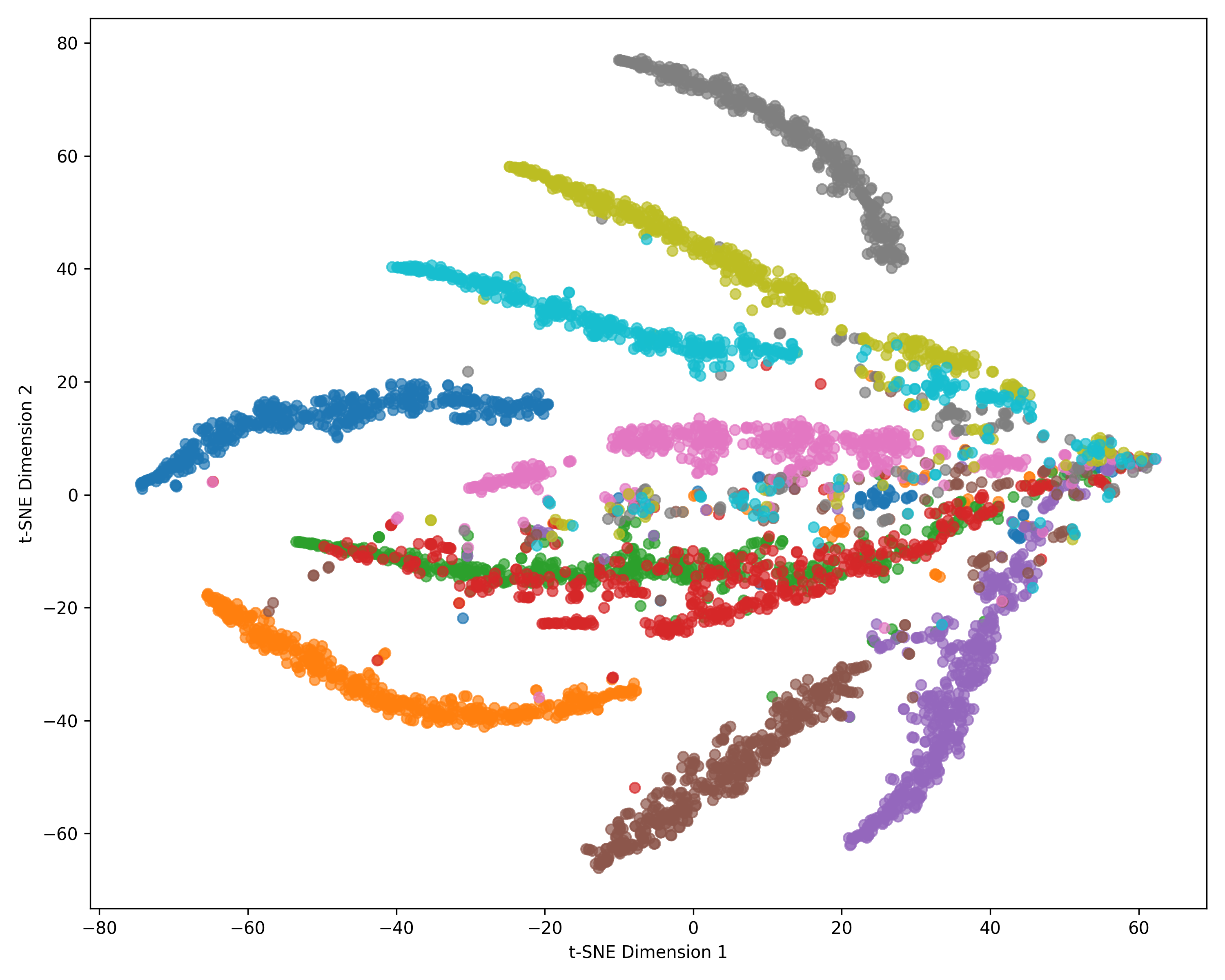}
        \caption{ER-AML: t-SNE visualization after training on task 20.}
        \label{fig:tsne_task1_after_task20_ER-AML}
    \end{subfigure}

    \vspace{0.5cm} 

    \begin{subfigure}[b]{0.48\textwidth}
        \centering
        \includegraphics[width=0.75\textwidth, height=2.5cm]{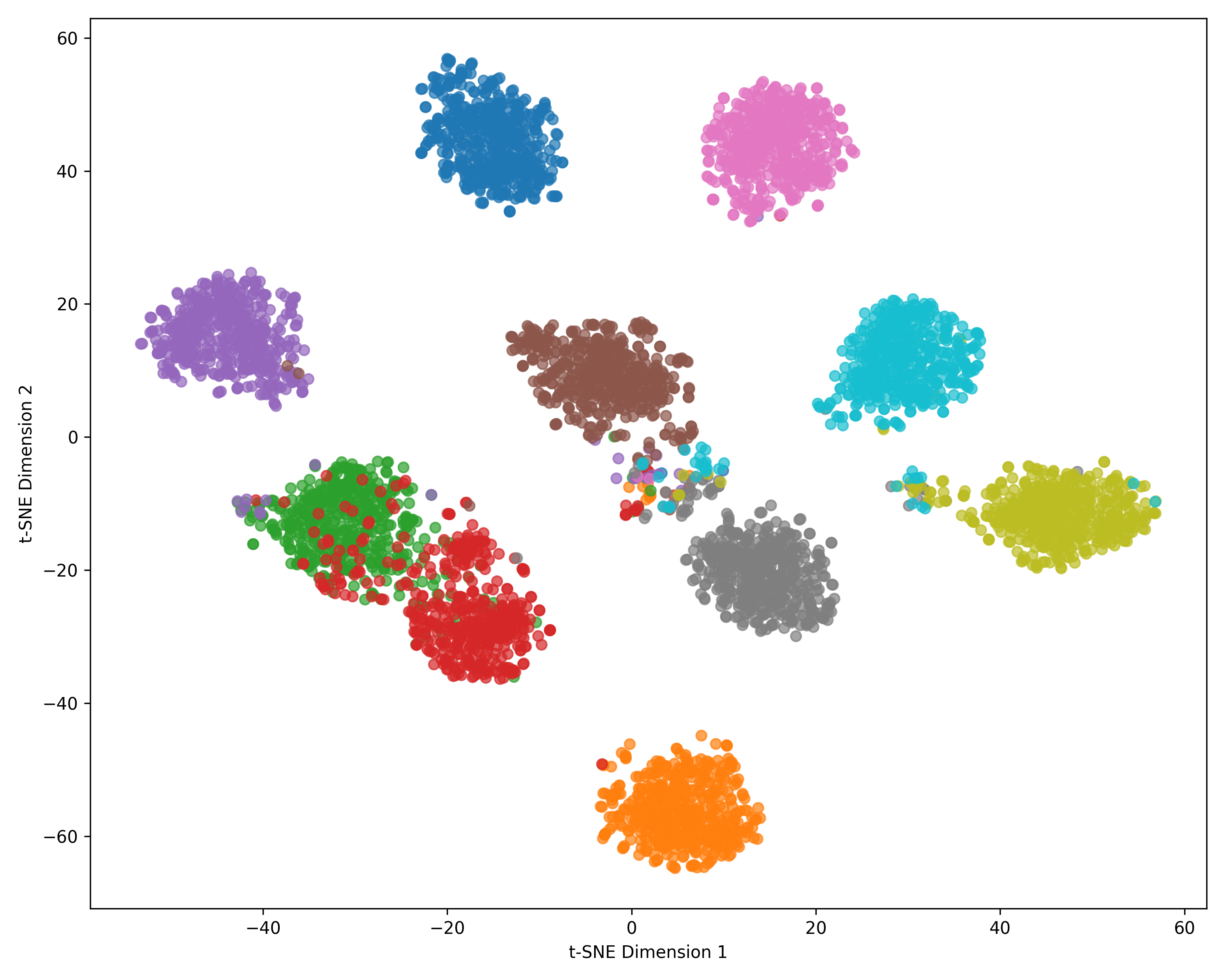}
        \caption{iSL-LRCP: t-SNE visualization after training on task 1.}
        \label{fig:tsne_task1_after_task1_proposed}
    \end{subfigure}
    \hfill
    \begin{subfigure}[b]{0.48\textwidth}
        \centering
        \includegraphics[width=0.75\textwidth, height=2.5cm]{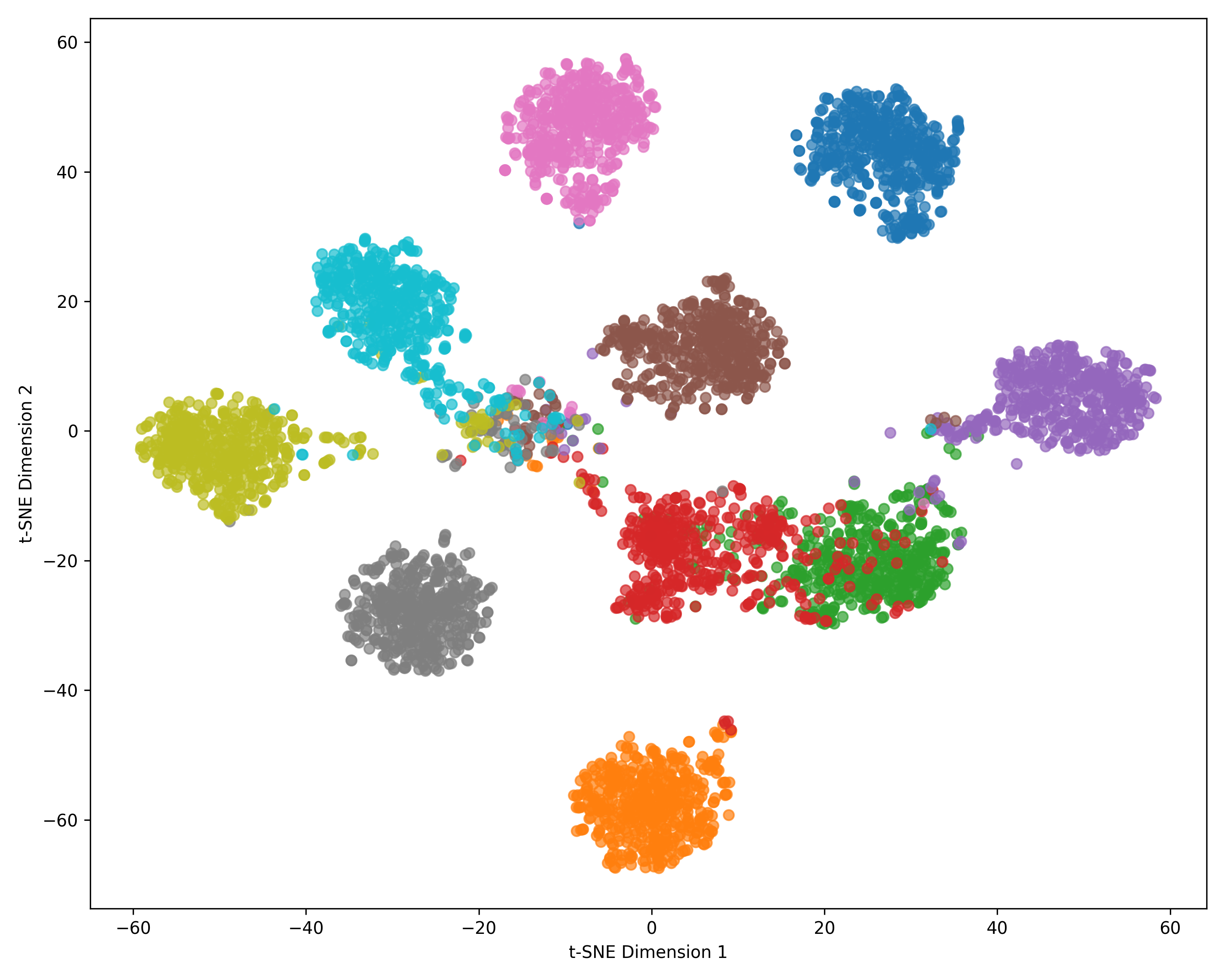}
        \caption{iSL-LRCP: t-SNE visualization after training on task 20.}
        \label{fig:tsne_task1_after_task20_proposed}
    \end{subfigure}

    \caption{t-SNE visualizations of the first task samples from SplitTinyImageNet under ER-AML and iSL-LRCP. Subplots (a) and (b) show the ER-AML results after training on task 1 and task 20, respectively. Subplots (c) and (d) depict iSL-LRCP results for the same conditions. The proposed method preserves cluster structures more effectively, demonstrating reduced deformation and overlap compared to ER-AML.}
    \label{fig:tsne_visualizations}
\end{figure*}

\subsection{Discussion}

\textbf{Ablations.} To evaluate the contribution of each loss component, we conducted an ablation study across three datasets, one for CI learning (sC100) and two for DI learning (R-MNIST and CORe50), as shown in Table \ref{table-loss-comparison}. The full loss function ($L_{\text{total}} = L_{\text{sc}} + L_{\text{push/pull}} + L_{\text{preserve}}$) consistently achieved the best results, highlighting the necessity of using cluster preservation and push-away/pull-toward mechanisms together. For instance, on sC100, the full loss function achieved 82.90\% accuracy, outperforming the combination without $L_{\text{push}}$ (82.48\%) or the use of only $L_{\text{push}}$ (20.70\%).

The cluster preservation loss ($L_{\text{preserve}}$) alone was sufficient to produce strong results, outperforming all baselines across the three datasets. For example, on CORe50, even with only $L_{\text{preserve}}$, the proposed method achieved 97.03\%, surpassing the offline baseline. This underscores the importance of the cluster preservation mechanism in maintaining latent space structures across tasks. However, incorporating push-away or pull-toward mechanisms with $L_{\text{preserve}}$ led to slight improvements, demonstrating their complementary role.

For DI settings, the pull-toward mechanism ($L_{\text{pull}}$) was used, although it requires the labels of the first task's class prototypes. While this approach deviates from the label-free replay paradigm, it helped achieve slightly better results. Nonetheless, $L_{\text{preserve}}$ alone was sufficient to surpass all baselines and even the offline learning method on CORe50.

The results without $L_{\text{preserve}}$ in Table \ref{table-loss-comparison} indicate a significant drop in accuracy for sC100. This low performance is attributed to the model's inability to maintain stable cluster structures over sequential tasks. The cluster preservation loss is critical for mitigating the adverse effects of distributional shifts in the latent space, thereby significantly reducing catastrophic forgetting. This observation underscores the necessity of $L_{\text{preserve}}$ in achieving robust continual learning performance.

\textbf{Visualizing Cluster Preservation.} Figure \ref{fig:tsne_visualizations} demonstrates the t-SNE plots \cite{vanDerMaaten2008} of the first task's samples from SplitTinyImageNet. Subplots (a) and (b) correspond to ER-AML, subplots (c) and (d) depict iSL-LRCP. For each method, the samples are shown after training on task 1 and task 20 (last task), respectively. The visualizations highlight that ER-AML results in significant cluster deformation and overlap after training on the final task, whereas iSL-LRCP method effectively preserves cluster structures with minimal overlap. This demonstrates the effectiveness of the cluster preservation loss in maintaining latent space representation. The reduced deformation and overlap directly correlate with lower forgetting and better retention of the prior knowledge.

\section{Conclusion}
This paper introduced a novel prototype-based continual learning framework that leverages a label-free replay buffer and cluster preservation loss to address catastrophic forgetting in both class-incremental and domain-incremental settings. By combining supervised and unsupervised contrastive losses with push-away and pull-toward mechanisms, the proposed method ensures effective retention of prior knowledge while adapting to new tasks. Experimental results on multiple benchmarks demonstrated superior performance compared to state-of-the-art baselines, highlighting the effectiveness of cluster preservation in maintaining structural consistency and enabling robust continual learning across diverse scenarios.
\section{Acknowledgement}

This work is supported by ELSA – European Lighthouse on Secure and Safe AI funded by the European Union under grant agreement No. 101070617.
{
    \small
    \bibliographystyle{ieeenat_fullname}
    \bibliography{main}
}


\end{document}